\documentclass[journal]{IEEEtran}
\ifCLASSINFOpdf
  \usepackage[pdftex]{graphicx}
  \usepackage{caption}
\else
\fi
\begin{document}

\title{Pedestrian Tracking with Gated Recurrent Units and Attention Mechanisms}
\author{Mahdi Elhousni \& Xinming Huang\\ 
Worcester Polytechnic Institute}

\maketitle

\begin{abstract}
Pedestrian tracking has long been considered an important problem, especially in security applications. Previously, many approaches have been proposed with various types of sensors. One popular method is Pedestrian Dead Reckoning (PDR) \cite{Beauregard2006PedestrianDR} which is based on the inertial measurement unit (IMU) sensor. However PDR is an integration and threshold based method, which suffers from accumulation errors and low accuracy. In this paper, we propose a novel method in which the sensor data is fed into a deep learning model to predict the displacements and orientations of the pedestrian. We also devise a new apparatus to collect and construct databases containing synchronized IMU sensor data and precise locations measured by a LIDAR. The preliminary results are promising, and we plan to push this forward by collecting more data and adapting the deep learning model for all general pedestrian motions.
\end{abstract}


%
\IEEEpeerreviewmaketitle

\section{Introduction}
%
%
%
%
\IEEEPARstart{T}{racking} a subject indoors is a very important challenge to solve. The first use case that comes to
our mind and that motivated this research was to be able to track in real time, firefighters and first
responders when they are intervening on a scene, using a lightweight and uncumbersome sensor. This
becomes more complicated to solve than the classical indoor pedestrian tracking when we consider
the different complex movements that they have to demonstrate in order be able to move in very
damaged areas. Classical tracking methods were tuned to work well when the motion is restricted
to straightforward walking. However, they fail and do not have the potential to be generalized
when other movement as common as running or as original as knee-crawling or belly-crawling are
demonstrated. On the other hand, a deep learning approach has this potential, because the tracking
accuracy and types of motions that it will be able to track are going to be only limited by the amount
and diversity of the data that it is being trained on. The way that data is being collected is also an
interesting topic. Typical databases containing Stride Lengths (SL) data are usually build using one of
the following methods :
\begin{itemize}
\item Placing labels on the ground, usually equally spaced, that the subject is supposed to step on
to follow a path. This usually produces a very limited range of SLs and does not give a true
representation of a free motion.
\item Using a treadmill with pressure sensors, or combining it with a foot pressure sensor. This only
produces straightforward walks with no change in orientation, making the data limited for
tracking purposes.
\item Using multiple synchronized cameras to track an object placed on the subject, in a restrained and
closed indoor space. The main issue with this method is the range of the camera limits the data
collection to very small space. Setting up such a setup can also be very costly.
\end{itemize}{}

We propose a new method to collect synchronized sensor and SL data by combining a LiDAR
and an IMU. Our method gives the subject the possibility to move freely in a big indoor space so that
we can produce data that will represent as close as possible a the true motions and displacements that
would happen in a free movement.

\section{Previous Work}

\subsection{Pedestrian tracking}
\subsubsection{Pedestrian Dead Reckoning (PDR)}
Dead Reckoning is a navigation method where one's position is predicted and calculated using it's precedent position, which is moved based on estimated speeds over a certain period of time. When applied to pedestrians, we obtain the main method used today to track people : Pedestrian Dead Reckoning \cite{Beauregard2006PedestrianDR}. Multiple variants of this method have been proposed, using different sensors and different body placements of the sensors used \cite{Beauregard2006AHP,Pratama2012SmartphonebasedPD,Kakiuchi2013PedestrianDR}, however the best results are obtained by using a foot-mounted IMU \cite{Jimnez2009ACO}. The PDR process follows a few main steps : divide the IMU data into 'steps' using a manually set threshold, derive the distance traveled during those steps by integrating the accelerometer data from IMU twice, derive the orientation of the subject by integrating the gyroscope data from the IMU twice, and finally combine the number of steps, distance traveled and orientations to re-construct the path of the pedestrian. \par
The accuracy of this method is extremely sensible to few variables such as the placement of the sensor on the foot and the selected threshold to detect steps. It also only works for a very limited range of movements : Straightforward walking. There is no way today to generalize PDR to more complex movements such as walking backwards, knee-crawling or belly-crawling, which are essential motions for firefighters and first responders for example.
\subsubsection{Machine Learning Approaches}
Predicting Stride Lengths (SL) using IMU data and deep learning has been attracting some attention in the last years. For medical purposes, Julius Hannink et al. proposed in \cite{article} to use Deep Convolutional Neural Networks (CNN) to extract multiple gait parameters such as the SL or the swing and stance time, using data collected from a foot-mounted IMU, in order to detect neurological and musculoskeletal diseases affects human gait quality. This work however does not aim to track the subject and assumes a straightforward motion on a treadmill with no turns. \par
On the other hand, Marcus Ede et al. have proposed in \cite{Edel2015AnAM} a B-LSTM network to predict the SL with the objective of abandoning the double integration methods in PDR and were able to obtain impressive results. However, in order to obtain ground truth, they placed labels along a path at 50 cm evenly spaced intervals and asked the subject to step on top of the labels, which is not an accurate representation of the motions present when a human is freely walking. They also did not adress the orientation issue when tracking a subject.
\subsection{Neural Networks}
\subsubsection{Recurrent Neural Networks (RNN)}
\begin{itemize}
\item	RNN : \par
Recurrent Neural Networks are a type of deep learning models that were first explored by John Hopfield during the 80's. They were designed to deal with sequential information, which makes them very suited to tasks such as handwriting recognition or speech recognition. An RNN consists of a hidden state and an optional output which operates on a variable or fixed length sequence. Here, contrary to traditional neural networks, the input and output are not independent from each other, because RNNs have a "memory" that lets them capture relevant information. Multiple variations of RNN have been proposed and tested such as Multiple Inputs RNN's, Long Short Term Memory (LSTM) or Gated Recurrent Units (GRU). \\
\item	GRU : \par
Gated Recurrent Units are type of RNN introduced by Cho et al. in \cite{Chung2014EmpiricalEO} which have the particularity of being able to retain information for a long time sequence. It is a gated mechanism where gating is done using the previous hidden state and the bias. GRUs contain two gates which are :
\begin{enumerate}
\item	Update Gate : 
\begin{equation}
g_u = \sigma(W_{ux}X_t + W_{uh}h_{t-1} + b)
\end{equation}
\item	Reset Gate : 
\begin{equation}
g_r = \sigma(W_{rx}X_t + W_{rh}h_{t-1} + b)
\end{equation}
\end{enumerate}
The hidden state is calculated as follows :
\begin{equation}
h_t = (1 - g_u).h_{t-1} + g_u.q_t
\end{equation}
with
\begin{equation}
q_t = \tanh(W_{hx}X_t + W_{hh}.(g_r.h_{t-1}) + b)
\end{equation}
that let us control what we want to keep from the previous state. \par


\quad \quad GRUs were proposed as an improvement to LSTMs, which are an older gated mechanism proposed by Sepp Hochreiter et al. in \cite{hochreiter1997long}. They were able to achieve comparable results in Natural Language Processing \cite{Kumar2015AskMA,Bansal2016AskTG} while being faster and consuming a lot less resources.
\end{itemize}

\subsubsection{Attention Mechanisms}
Attention Mechanisms are a popular trend nowadays in Image Recognition, Neural Machine Translation and Speech Recognition tasks \cite{Xu2015ShowAA,NIPS2017_7181,NIPS2015_5847}. This mechanism allows for a more straightforward dependence between the state of the model at different points during the time sequence. It basically helps the model to "pay attention" to the most important data in the sequence. We follow the formulation proposed by Colin Raffel et al. in \cite{Raffel2015FeedForwardNW} where the "context" vector for the entire sequence $c$ is defined as:
\begin{equation}
c_t = \sum_{t=1}^{T} \alpha_th_t
\end{equation}
where $T$ is the total number of time steps in the input sequence.
The weightings  $\alpha_{t}$ can be computed by:
\begin{equation}
e_t = a(h_t) \quad , \quad
\alpha_t = \frac{\exp(e_t)}{\sum_{k=1}^{T} \exp(e_k)}
\end{equation}

 \par
As we can see, the hidden state $h_t$ is passed through a learnable function $a$ that depends on $h_{t}$ in order to produce a probability vector $\alpha$. The context vector is then computed by weighting $h_t$ with $\alpha$ and is then used to compute a new state sequence $s$ where  $s_{t}$ depends also on the hidden state  $h_t$. This formulation is called "Feed-Forward Attention" and it was shown that it is very efficient and performs very well when solving long-term memory problems for long sequences. This is adapted to our task because sensor data received at high frequency tends to form very long time sequences.

\section{Problem Reformulation}
In order to track the trajectory of a pedestrian, we need to have both the SL and the orientation at
a given moment in time. 
We propose, in the context of a 2D trajectory in the $(x,y)$ plan, to decompose our prediction of
the displacement d into 2 predictions: $dx$ and $dy$. By predicting $dx$ and dy instead of $d$ only, we are
implicitly predicting the subjects orientation too. This method is not limited to 2D movements only,
since the elevation dz can be obtained using other sensors, and thus form the full 3D trajectory in the
$(x,y,z)$ plan.

\section{Data Collection}
In order to collect our ground truth and construct our database, we combined a lightweight
foot-strapped IMU with a 2D LiDAR. More details below.
\subsection{Hardware}
\begin{itemize}
\item NGIMU ; The NGIMU is an IMU released by x-io Technologies : It was chosen because it is lightweight,
compact and easily configurable. The sensors included in this unit are a triple-axis gyroscope,
accelerometer, and magnetometer, as well a barometric pressure sensor and humidity sensor.
The communication between the IMU and the computer was established using the OSC protocol
library in python, through the Robotic Operating System (ROS) system. The sensor data is time
stamped on-board when registered by the sensor, making synchronization with others sensor
possible. The IMU was strapped to the foot of the subject when recording data with the z axis
pointing upward and the x axis pointing forward.
\item Hokuyo 2D LiDAR ; In order to register the displacement of the subject, we used a 2D Hokuyo LiDAR, namely the
"Hokuyo UXM-30LAH-EWA Scanning Laser Rangefinder". This LiDAR was chosen because of it
high range of 80 meters. This gives us the possibility to track our subject over long distances
while walking around in a room. It has a scan angle of 190 $^\circ$ and an average accuracy of ± 30mm.
It outputs through an Ethernet 100Base-TX interface, and we used the Hokuyo ROS driver to
obtain time-stamped samples.
\end{itemize}
\subsection{Method}
In order to register the displacement of the subject when walking indoors, we use a 2D LiDAR.
We start by scanning the room to get a reference frame $F_0$ , before allowing the subject to start walking
around with the IMU strapped to their foot. To keep track of the position of the subject, we compare
the incoming frames $F_i$ from the LiDAR with the reference frame $F_0$ to see what are the points that
have moved. The centroid of the points that we obtain after subtracting $F_i$ from $F_0$ represent the subjects position. We
then define a constant time period of 2 seconds, and calculate the displacement during these periods
by using our registered centroids positions. \\
We recorded a total amount of 90 minutes of an adult pedestrian walking freely at a normal pace.
We used a 250 $hz$ frequency for the IMU which amounts to more than 1 000 000 samples of the triple
axis gyroscope, accelerometer and magnetometer. We also collected data for other motions such as running and standing still. However, we decided to concentrate on the
walking part first, and leave multi-motions tracking for future work. 
\subsection{Synchronization}
First, the data collected is aligned based on the timestamps attached to each sample. However, during our experiements, we noticed that the LiDAR data still suffer 
from a small delay.
To synchronize the data further, we use a signal alignment
technique based on the fact that integrating the accelerometer data once during a very small period
of time guarantees that they will be no significant drift : We start by recording multiple 2 or 3 seconds
of the IMU and LiDAR data where we try to produce a single spike by moving the IMU once in the field of view of the LiDAR. We then use the IMU data and LiDAR data to generate 2 velocity graphs $V_i$
and $V_l$ and use the spikes in both signals to align them. After calculating the delay value in all the
recorded samples, we are able to calculate an average value of what the delay between the two sensors
is, which is 3.89 ms. Fig. 1 shows one of the velocity plots used to calculate the delay average between the IMU and LiDAR.

\begin{figure}[h]
\centering
\captionsetup{justification=centering}
\includegraphics[height=35mm,width=70mm]{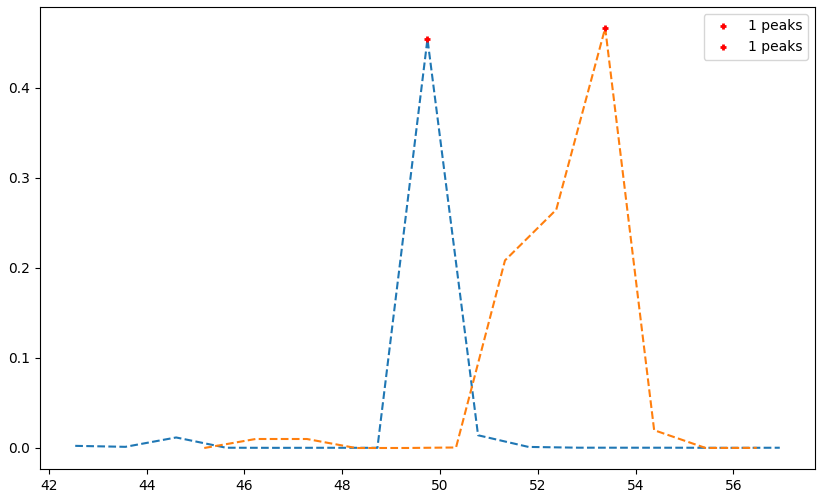}
\caption{Example of one of the velocity graphs used to calculate the delay average}
\end{figure}

\section{Stride Lengths Predictions}

\subsection{Preparing and Segmenting the Data}
We recorded the data in runs of 10 or 5 minutes. We always start by clipping the 3 first and last
seconds of each run, which usually correspond to the periods when the subject was getting in or out of
the frame of the LiDAR. We then proceed to normalize the data between $[-1,1]$ and use a sliding
window to segment it in sequences of 2 seconds each. After aligning the LiDAR and IMU data, we use
the time stamps to assign two label dx and dy to each of the previously segmented sequences. The final
data contained in the segmented sequences is a concatenation of the data generated by the triple-axis
accelerometer, gyroscope, magnetometer and their magnitudes.

\section{Neural Network}

Our network structure consists of two stacked
GRU layers \cite{Chung2014EmpiricalEO} , followed by a drop out layer with a
ratio of 0.25. We add on top of that an Attention
layer \cite{Raffel2015FeedForwardNW} and finally 2 dense layers, with the last
layer being the final regression one with a linear
activation function.
We used the same network architecture to
predict both $dx$ and $dy$. Both networks were
implemented using Keras with a TensorFlow
backend, and trained on an Nvidia Tesla K40. The
networks were trained for 60 epochs, using a batch
size of 5, the Mean Absolute Error (MAE) as a
loss function and an RMSProp optimizer with a
strarting learning rate of 0.001 that was decreased
by a factor of 0.2 whenever the validation loss
stopped improving for more than 10 epochs. The
input dataset set was split into a training and
validation datasets using a 0.8 split ratio. The
testing dataset was not included in the input
dataset. We also trained multiple variants of our
network to prove the usefulness of the Attention
layer. The other trained network (namely : Bilinear-LSTM \cite{Edel2015AnAM},
GRU, 2GRU, 3GRU) followed the same training
procedure and used the same data.

Table 1 shows the order and details of the different layer
that were used to build our Neural Network.

\begin{table}[h]
	\centering
    \caption{Neural Networl Layers}
    \begin{tabular}{||c|c|c||}
      \hline
       \textbf{Layer} & \textbf{Input} & \textbf{Output}\\
      \hline     
      $Input Layer$ & (500,12) & (500,12)\\
      \hline
      $GRU_1$ & (500,12) & (500,256)\\
      \hline
      $GRU_2$ & (500,256) & (500,256)\\
      \hline
      $Drop  Out$ & (500,256) & (500,256)\\
      \hline
      $Attention Layer$ & (500,256) & (1,256)\\
      \hline
      $Dense Layer$ & (1,256) & (1,64)\\
      \hline
      $Dense Layer$ & (1,64) & (1)\\
      \hline
    \end{tabular}
\end{table}

\section{Results}
When it comes to predicitng SL for a freely moving human subject, our tests show that the proposed architecture outperforms any other variant or previously
proposed network. Also the attention layer seems to make the training more stable and makes it
converge faster. Our models are also very lightweight : only 8mb, and requires only 82ms to produce a
prediction, making them ideal for real time deployment on embedded structures. \par Fig. 2 and Fig. 3 list the different MAE scores of the different networks that were trained. Our method (2GRU+ATT) is shown to have the smallest final error value on both the test and validation datasets. Our final MAE for both the $dx$ and $dy$ models are $0.19$ and $0.20$ respectively for the validation
dataset, and $0.25$ and $0.24$ respectively for the testing dataset.
Fig. 3 shows a qualitative comparaison of some of the plots
generated after feeding the test dataset to our trained network. We are able to reproduce the path
followed by the subject by combining the predictions of both the dx and dy models.

\begin{figure}[h]
\captionsetup{justification=centering}
\centering
\includegraphics[height=45mm,width=70mm]{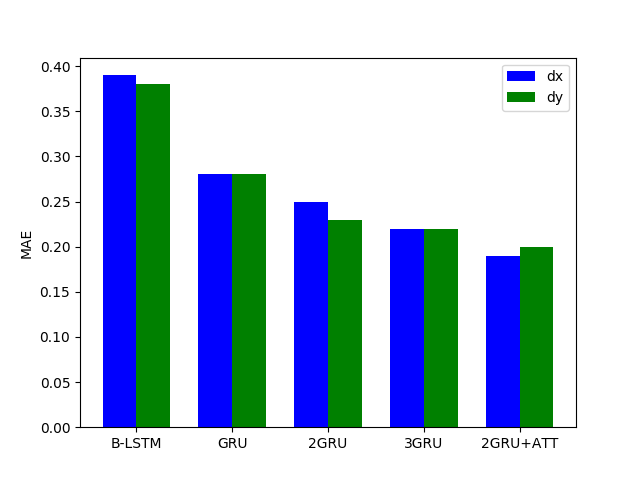}
\caption{MAE Scores on the Validation Dataset.}
\includegraphics[height=45mm,width=70mm]{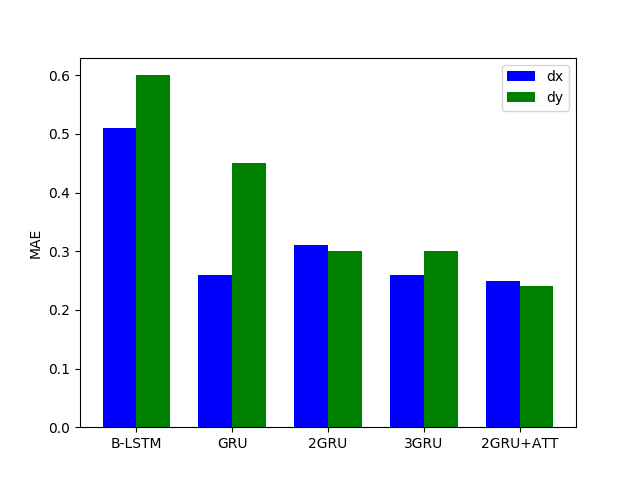}
\caption{MAE Scores on the Testing Dataset.}
\end{figure}

\begin{figure}[h]
\captionsetup{justification=centering}
\includegraphics[height=30mm,width=41mm]{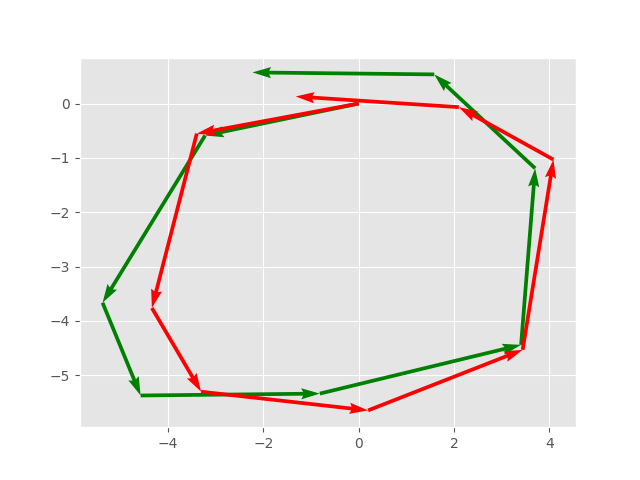}
\includegraphics[height=30mm,width=41mm]{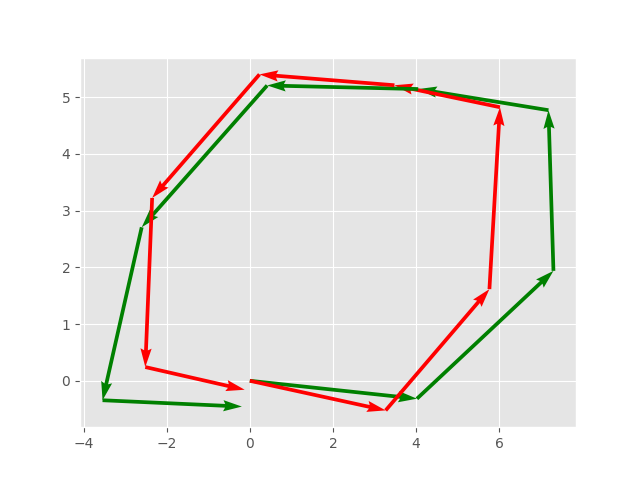}
\includegraphics[height=30mm,width=41mm]{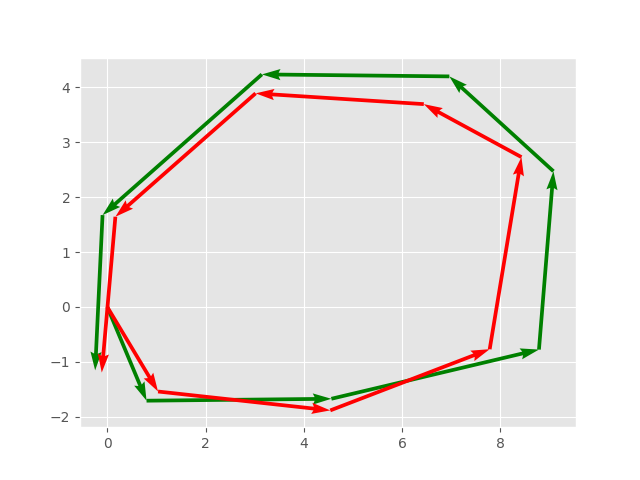}
\includegraphics[height=30mm,width=41mm]{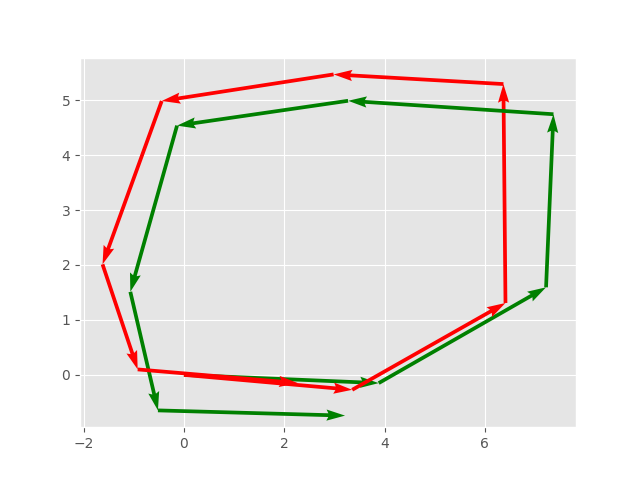}
\includegraphics[height=30mm,width=41mm]{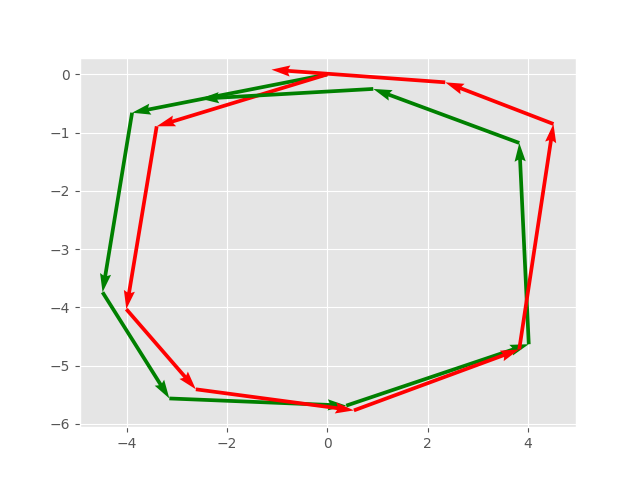}
\hspace{0.5cm}
\includegraphics[height=30mm,width=41mm]{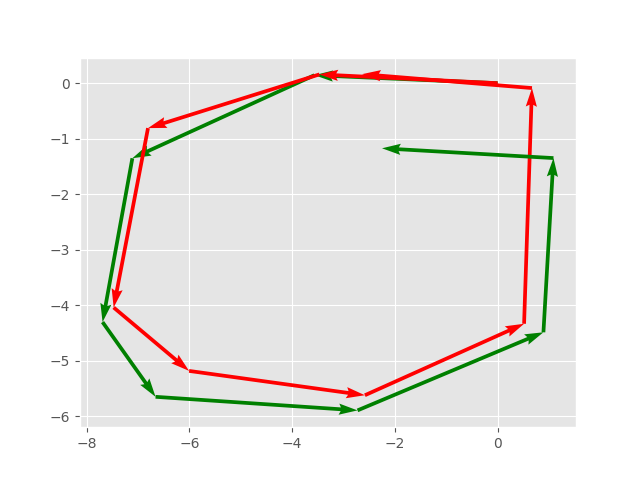}
\caption{Qualitative results on the Test Dataset.
Red is the groundtruth while green is the combined predictions.}
\end{figure}

\textbf{Ablation Study} : 
Fig. 5 and Fig. 6 show a comparison between the training plots of the NN with and without the Attention Layer for both the $dx$ and $dy$ networks. We can see that the Attention Layer helps the network to be more stable while learning and makes it converge faster.

\begin{figure}[h]
\centering
\captionsetup{justification=centering}
\includegraphics[height=40mm,width=80mm]{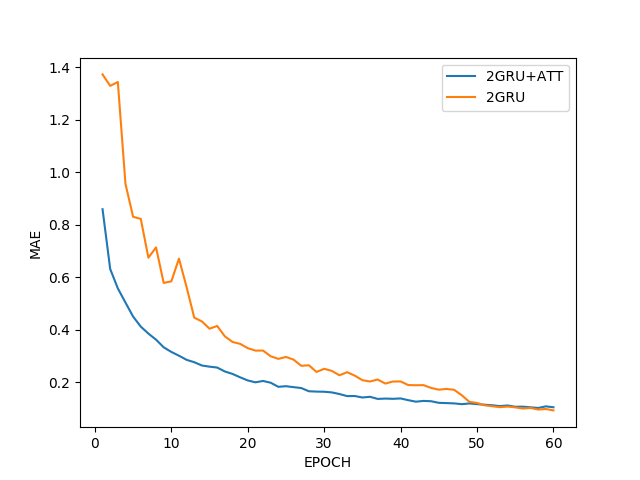}
\caption{Evolution of the MAE during training for $dx$}
\includegraphics[height=40mm,width=80mm]{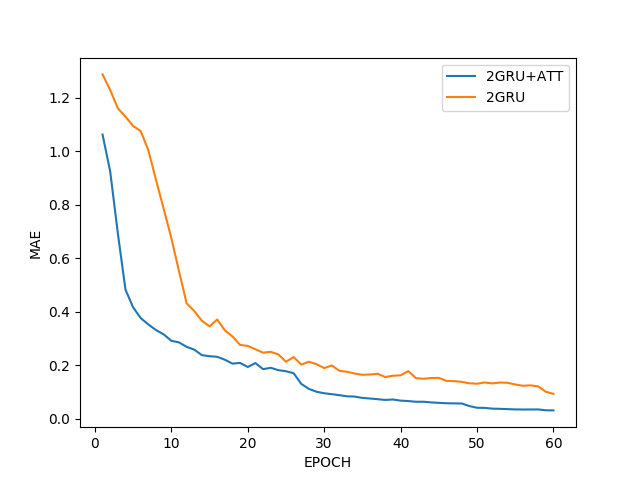}
\caption{Evolution of the MAE during training for $dy$}
\end{figure}

\section{Conclusion}
The initial experimental results shows the proposed method is promising. As more data are collected for training, the prediction model has the potential to become more accurate. In its present form, it provides good estimates of dx and dy displacement when the subject moves straight. But when the subject is turning, the predictions become less accurate. This is largely owing to the fact that the turning segments in our training dataset are disproportionally less than the straight movement. In the future work, we plan to collect more data and to generalize this approach to all typical motions such as walking, running, crawling, etc. The goal is to develop a generalized deep learning model that can track pedestrians even with complex motions.


%

\ifCLASSOPTIONcaptionsoff
  \newpage
\fi



\bibliographystyle{IEEEtran}
\bibliography{bibfile1.bib}
%

%

\end{document}